\documentclass[a4paper,journal]{IEEEtran}
\usepackage{amsmath,amsfonts}
\usepackage{algorithmic}
\usepackage{algorithm}
\usepackage{array}
\usepackage[caption=false,font=normalsize,labelfont=sf,textfont=sf]{subfig}
\usepackage{textcomp}
\usepackage{stfloats}
\usepackage{url}
\usepackage{verbatim}
\usepackage{graphicx}
\usepackage{cite}
\usepackage{multirow}
\usepackage{bbding}
\usepackage{hyperref}
\hypersetup{hidelinks, colorlinks=false, pdfstartview=Fit, breaklinks=true}

\begin{document}

\title{Successive optimization of optics and post-processing with differentiable coherent PSF operator and field information}

\author{Zheng Ren, Jingwen Zhou, Wenguan Zhang, Jiapu Yan, Bingkun Chen, Huajun Feng, Shiqi Chen
\thanks{\IEEEcompsocthanksitem Zheng Ren, Jingwen Zhou, Wenguan Zhang, Jiapu Yan, Bingkun Chen, Huajun Feng, Shiqi Chen are with the State Key Laboratory of Extreme Photonics and Instrumentation, College of Optical Science and Engineering, Zhejiang University, Hangzhou, 310027, China. 
\IEEEcompsocthanksitem Corresponding author: Shiqi Chen (E-mail: chenshiqi@zju.edu.cn)
\IEEEcompsocthanksitem This work is supported by National Natural Science Foundation of China (No. 62275229). 
}
\thanks{Manuscript received Sept 17, 2024.}}



\maketitle

\begin{abstract}
Recently, the joint design of optical systems and downstream algorithms is showing significant potential. However, existing rays-described methods are limited to optimizing geometric degradation, making it difficult to fully represent the optical characteristics of complex, miniaturized lenses constrained by wavefront aberration or diffraction effects. In this work, we introduce a precise optical simulation model, and every operation in pipeline is differentiable. This model employs a novel initial value strategy to enhance the reliability of intersection calculation on high aspherics. Moreover, it utilizes a differential operator to reduce memory consumption during coherent point spread function calculations. To efficiently address various degradation, we design a joint optimization procedure that leverages field information. Guided by a general restoration network, the proposed method not only enhances the image quality, but also successively improves the optical performance across multiple lenses that are already in professional level. This joint optimization pipeline offers innovative insights into the practical design of sophisticated optical systems and post-processing algorithms. The source code will be made publicly available at \url{https://github.com/Zrr-ZJU/Successive-optimization}
\end{abstract}


\section{Introduction}
\IEEEPARstart{W}{ith} the rise of mobile photography, traditional lens design for smartphones is increasingly pushing towards optical limits, striving to balance high imaging quality with limited module space. To compensate for inherent system flaws (\textit{e.g.}, aberration, glare, manufacturing errors), numerous image post-processing algorithms have been developed\cite{Chung:19,Chen_2023,lin2022non,li2021universalflexibleopticalaberration,Chen_2021,eboli2022fasttwostepblindoptical}. In recent years, to fully harness the potential of these two distinct design stages, there has been significant development of joint optimization pipelines that effectively integrate optical modeling with image post-processing algorithms \cite{Hale:21,Sun2021DiffLens,dO,Yang_2024,Li:21,cote2023differentiable}. 

The joint optimization paradigm has been successfully applied to the design of simple optical systems, such as diffractive optical elements (DOEs), using a differentiable paraxial Fourier image formation model \cite{Dun:20,DeepOpticsDepth,DeepOpticsHDR,LearnedOpticHDR,EndToEndCam}. To tackle the more complex compound lens systems found in commercial cameras, Tseng et al. \cite{Tseng2021DeepCompoundOptics} developed a proxy model to generate the point spread function (PSF), while other works \cite{dO,Yang_2024,cote2023differentiable,nie2023freeform} have directly implemented differentiable ray-tracing operations within automatic differentiation frameworks \cite{pytorch,tensorflow}. This joint design approach shifts part of the high-quality imaging burden onto neural networks, reducing the emphasis on sensor image quality and providing greater flexibility in optical design\cite{zhou2024revealing}. 

Current open-source ray-based methods often ignore the wave characteristics of light, calculating the PSF incoherently using geometric interpolation \cite{dO,Yang_2024} or Gaussian approximation \cite{Li:21,cote2023differentiable} This limitation restricts joint design to lens scenarios where geometric aberrations dominate. However, in advanced compact mobile modules (with pixel sizes often less than 2 µm), the PSF is primarily influenced by wavefront aberrations and aperture diffraction effects, rendering these incoherent PSF calculation methods inadequate. Consequently, previous works have focused on finding a good initial configuration \cite{Yang_2024,gao2024globalsearchopticsautomatically} or a low-cost solution \cite{cote2023differentiable} for cameras used in computer vision tasks, without fully exploring the potential of joint optimization in lens optical design. In contrast, a refined and accurate coherent PSF calculation provides a promising avenue for deeper exploration of both cameras and post-processing algorithms in joint optimization.

Differentiable ray tracing demands substantial computational memory for joint optimization. This challenge is mitigated by various methods, such as simplifying the intersection solver \cite{dO}, bridging image gradients between two stages \cite{Yang_2024,gao2024globalsearchopticsautomatically}, or employing a low sampling rate \cite{Sun2021DiffLens}. Further memory-efficient approaches could be developed by exploring alternative perspectives.

\begin{figure*}[ht]
    \centering
    \includegraphics[width=0.9 \textwidth]{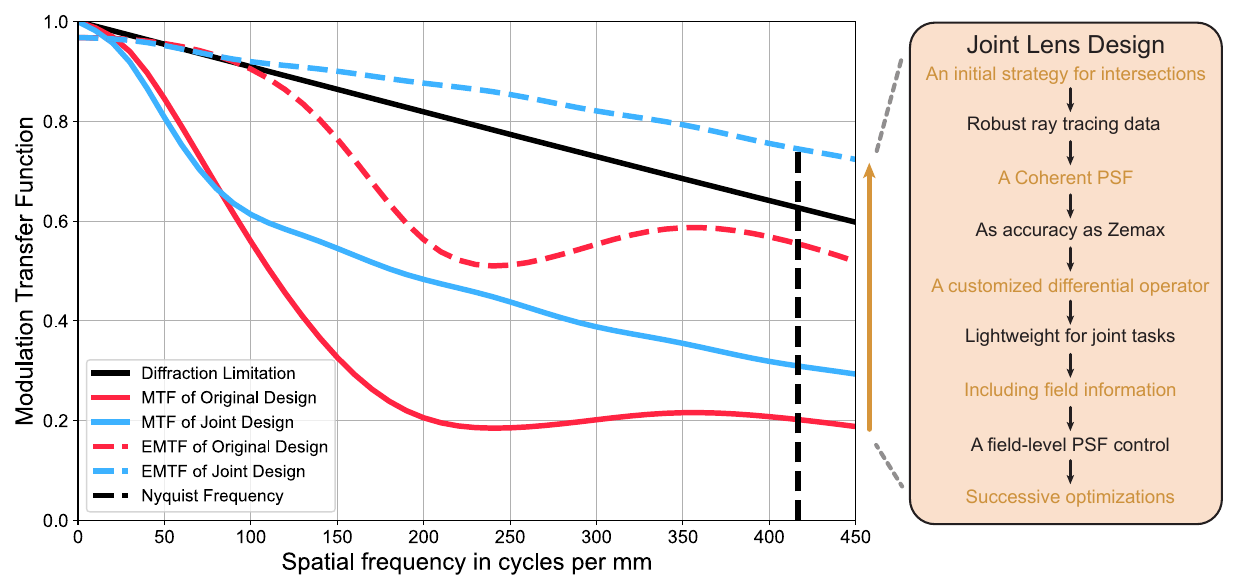}
    \caption{Schematic diagram of this work. In the left figure, we focus on successively optimizing the optical and image performance of advanced complex lens using a joint approach. Compared to the existing joint design paradigm, our key features are highlighted in the right flowchart.}
    \label{fig:intro}
\end{figure*}

In this work, we introduce an efficient and differentiable optical simulation model that is capable of simulating and differentiating through complex refractive lenses. This model accounts for highly aspherical surfaces, wavefront aberrations, and aperture diffraction effects. A novel initial value strategy within Newton's method is employed to improve the accuracy of inner intersection points on highly aspherical surfaces. Using reliable ray tracing data, our coherent PSF consistently matches the Huygens PSF results from Zemax across a wide range of lens scenarios. Additionally, by manually back-propagating (BP) gradients through the differential operator, we successfully decoupled the ray and grid dimensions in coherent PSF calculations, leading to an 18.4-fold reduction in memory usage.

\begin{figure*}[ht]
    \centering
    \includegraphics[width=1\textwidth]{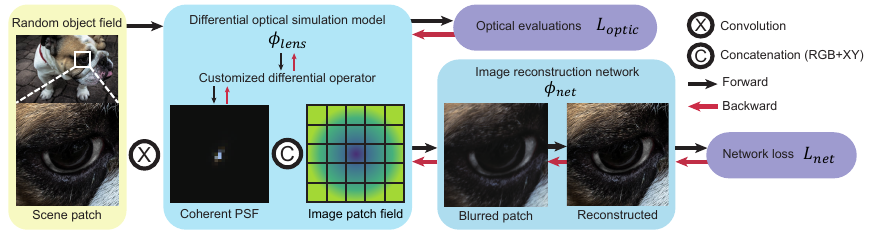}
    \caption{Overview of our joint optimization pipeline. In each iteration, a random object field is sampled to construct the input for the downstream restoration network, consisting of the blurred image patch and the normalized pixel coordinates. The lens parameters $\phi_{lens}$ are optimized based on the PSF loss derived from reconstruction errors and on optical evaluations, which consider both optical performance and geometrical constraints through exact ray tracing. Concurrently, the reconstruction network parameters $\phi_{net}$ are trained to minimize the reconstruction loss while adapting to the spatially varying lens aberrations.}
    \label{fig:joint}
\end{figure*}

Field information is crucial to the joint optimization pipeline, enabling a novel and comprehensive evaluation of spatially varying PSF in advanced optical design, which leads to the reconstruction of high-quality images. To achieve the highest possible optical and image performance, as shown in Fig.\ref{fig:joint}, we develop a joint optimization pipeline that leverages field information to address optical degradation related to the field-of-view (FoV). Various optical constraints are introduced to ensure the stability of the lens design specifications. Beyond producing high-quality images, the restoration network offers a new approach to overall PSF evaluation for optical optimization, progressively pushing lens performance closer to the diffraction limit.

Our contribution can be summarized as follows:
\begin{itemize}
\item Using a \textbf{customized differential operator} that efficiently arrange the memory, we develop a differentiable optical simulation model that avoids exponentially growing computation overhead and could accurately calculates coherent PSFs. 
\item A joint optimization pipeline is presented that \textbf{not only enhances image quality}, but also \textbf{successively improves the performance of optics} across multiple lenses that are already in professional level.
\item For the first time, we show that joint optimization could realize a \textbf{field-level PSF control} in advanced optical design, revealing its tremendous potential by bringing evaluated lenses approaching the diffraction limit with an improved effective modulation transfer function (EMTF).
\end{itemize}
In addition, we release our code in the hope of enabling further joint design applications.

\section{Differentiable Optical Simulation Model}
We utilize a ray-tracing based model for PSF formation, which can realize precise and differentiable results. The initial Rays are uniformly sampled on the entrance pupil according to the vignetting coefficient. And the subsequent The ray tracing process involves two sequential steps for each surface: 1) solving for the intersection point between the incident ray and the surface, and 2) updating the direction cosines of the refracted ray according to Snell's law. During propagation, rays undergo three validity checks: confirming the intersection solution, ensuring the intersection occurs within the surface's aperture, and verifying that no total internal reflection occurs. Additionally, we record all relevant data during ray propagation to compute the optical constraint terms in the loss functions.

To ensure the accuracy of Newton's method when solving for inner intersection points on high aspherics, we propose an effective initial value estimation strategy. Additionally, we use a coherent PSF computation method to model optical degradation caused by wavefront aberrations and diffraction effects. Furthermore, we have develop a differential operator with manual BP to reduce the memory cost associated with broadcasting tensors.

\subsection{Initial Guess for Intersections}\label{sec:initial}
\begin{figure}[ht]
    \centering
    \includegraphics[width=0.5\textwidth]{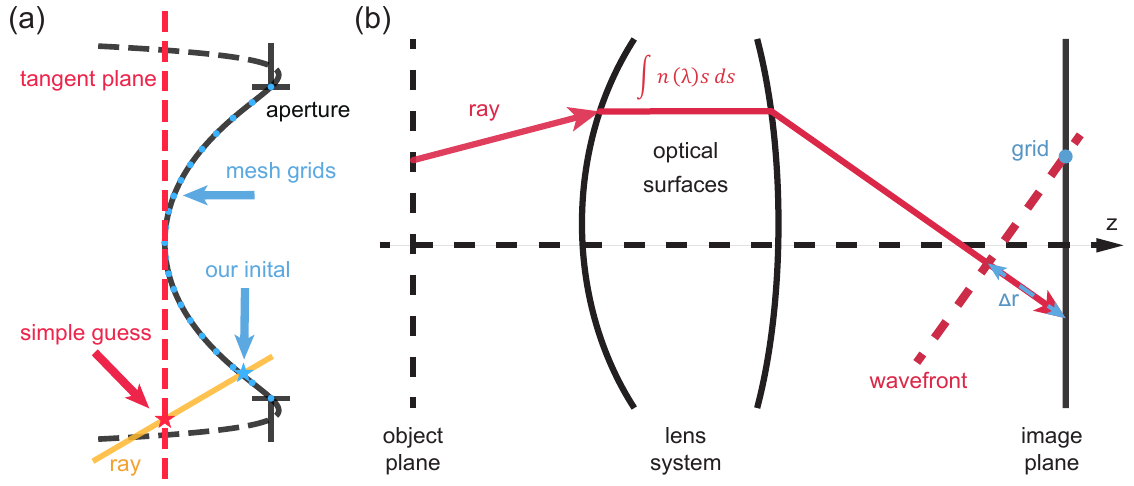}
    \caption{(a) shows the differences between our proposed initial value estimation strategy (blue) and the simple estimation method (red), with initial points marked by stars. (b) provides a schematic diagram of the coherent PSF calculation process.  Note that the spacing between the grid points and the ray trace points is exaggerated for visualization purposes.}
    \label{fig:schematic}
\end{figure}

In previous differentiable ray tracing models \cite{dO,Yang_2024,nie2023freeform}, Newton's method is employed iteratively to determine the intersection between rays and surfaces. Typically, the initial guess for Newton's iteration is set as the intersection of the ray with the tangent plane of the surface. However, highly aspherical surfaces can produce inflection points outside the aperture, potentially causing Newton's method to converge to intersection points outside the surface aperture using the previous simple initial guess strategy. To address this, we propose an initial value scheme that starts Newton's method iteration closer to the correct intersection point within the aperture. Specifically, we place reference points $P_{s}$ on the surface within the aperture. For each ray, we identify the reference point nearest to it in Euclidean distance:
\begin{equation}\label{eq: distance}
   P_{ref} = \arg\min_{P_{s}}\left(\frac{\left(P_{s}-P_{r} \right)\times D_{r}}{\left(P_{s}-P_{r} \right)\cdot D_{r}} \right),
\end{equation}
where $P_{r}$ and $D_{r}$ represent starting point and normalized direction of the ray, respectively. We propagate the ray to the perpendicular point $P_{initial}$ from the reference point to the ray, using $P_{initial}$ as the initial value for solving the intersection with Newton's method.
\begin{equation}\label{eq: prop}
   P_{initial}=P_{r}+\left(\left(P_{ref}-P_{r}\right)\cdot D_{ray}\right)D_{ray}.
\end{equation}
The visualized comparisons are shown in Fig. \ref{fig:schematic}(a). After Newton's method iterations, our proposed initial value scheme successfully identifies the correct intersection point within the aperture. In contrast, the initial value estimated from the tangent plane tends to converge to an incorrect intersection point outside the aperture.

\subsection{differentiable coherent PSF Operator}
When the optical path difference (OPD) of the system's exit pupil counts, geometrical estimation methods cannot accurately describe the PSF constrained by wave aberrations. Incoherently summing the intensity distributions of rays may underestimate the PSF convergence due to significant spot spreading and neglected diffraction effects. Coherent PSF calculation methods, such as fast fourier transform (FFT) and Huygens' wavelet theory, are widely used in commercial optical design software. These methods account for the phase differences between rays and compute the complex amplitude distribution on the image plane, with the PSF being the square of the magnitude of this distribution.

We developed a differentiable coherent PSF model in which rays are coherently summed as plane waves to form the complex amplitude. Each ray represents a plane wave originating from a field point, interacting with the optical system, and propagating to the image plane. We accumulate the optical path length of each ray from the field point to the image plane
\begin{equation}\label{eq: opl}
   OPL=\int_{field\ point}^{image\ plane} n \left( \lambda \right)s ds,
\end{equation}
where we use dispersion models consistent with Zemax material catlalogs (e.g., Schott, Sellmeier, etc.) to retrieve the refractive index at any wavelength $\lambda$. 

For the sampled grid points on the image plane, the complex amplitude distribution is obtained by coherently summing the complex field contributions of plane waves represented by each ray
\begin{equation}\label{eq: coherent}
   A\left( x,y \right)=\sum_{i} a_{i}e^{ik\left( OPL_{i}+\Delta r_{i}\left( x,y \right) \right)}\langle \vec {n},\vec {D} \rangle,
\end{equation}
where $A\left( x,y \right)$ is the complex amplitude at the grid point $\left( x,y \right)$, $a_{i}$ is the amplitude of the $i$-th ray, $k$ is the wave number, $\Delta r_{i}\left( x,y \right)$ denotes the optical path length of the $i$-th ray to the grid point $\left( x,y \right)$, and $\langle \vec {n},\vec {D} \rangle$ is the inner product of image plane normal and the cosine of the ray direction. The PSF represents the intensity captured on the detector, is calculated by
\begin{equation}\label{eq: psf}
   PSF\left( x,y \right)=A\left( x,y \right)A^{*}\left( x,y \right).
\end{equation} 
Fig. \ref{fig:schematic}(b) illustrates the trajectory of a single ray as it travels from the field point to the image plane, and the process of computing the complex field at grid points based on the accumulated optical path length and the optical path difference at the image plane.

\begin{figure}[ht]
    \centering
    \includegraphics[width=0.5\textwidth]{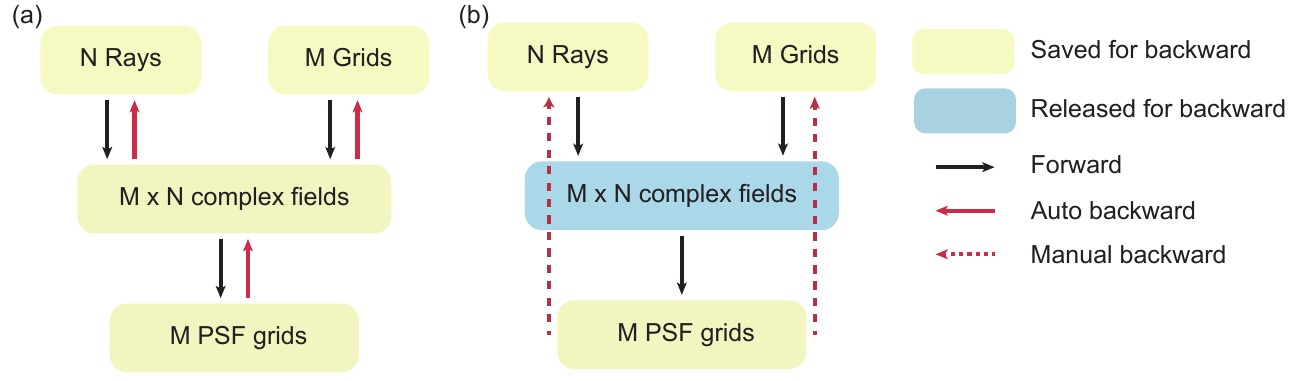}
    \caption{Comparison of the gradient computation flow. (a) demonstrates the normal calculation procedure in automatic differentiation functions.  (b) shows that the proposed differential operator manually back propagates the analytical gradients without saving the large broadcast tensors.}
    \label{fig:opetator diagram}
\end{figure}

Unlike incoherent PSF calculation methods, coherent PSF calculation involves summing the complex amplitudes of plane waves represented by each ray at every grid point. As illustrated in Fig. \ref{fig:opetator diagram}, directly applying typical automatic differentiation functions to compute the coherent PSF for N rays and M sampled grids results in broadcasting and generates intermediate variables of size MxN, which leads to excessive memory usage.

To address this issue, we propose a differential operator that calculate coherent PSF grids with only the original N rays and M grids saved for manual BP, as depicted in Fig.\ref{fig:opetator diagram}. During the backward pass, we compute the analytical gradients of the Eq. (\ref{eq: coherent}-\ref{eq: psf}) with respect to various variables using the chain rule. Calculating the derivative of the phase term $\varphi$ in Eq.(\ref{eq: coherent}) from real to complex numbers is relatively challenging:
\begin{equation}\label{eq: dpsfdphase}
\begin{split}
\frac{\partial PSF}{\partial \varphi} &= \frac{\partial PSF}{\partial \Re(A)}\frac{\partial \Re(A)}{\partial \varphi}+\frac{\partial PSF}{\partial \Im(A)}\frac{\partial \Im(A)}{\partial \varphi}\\
&=2\Re(A)(-\Im(A))+ 2\Im(A)\Re(A),
\end{split}
\end{equation}
where $\Re$ and $\Im$denote the real and imaginary parts of complex numbers, respectively. The remaining differential relations are all linear equations that are straightforward to derive. 
 
\subsection{Optical Image Formation} \label{sec: degrade}
To fairly and conveniently compare the performance of the jointly optimized optical system with the original optical design, we use three design wavelengths (486.1 nm, 587.6 nm, and 656.3 nm) to represent the three-channel $PSF_{c}$. The chief ray of the reference wavelength (587.6 nm) is designated as the center point of $PSF_{c}$, allowing us to model the longitudinal chromatic aberration.

Our differentiable lens simulation model generates the degraded image by applying a spatially varying convolution to the scene image $I_{s}$ and adding white Gaussian noise $\mathcal{N}$ with a standard deviation of $\sigma = 0.03$ to simulate sensor noise
\begin{equation}\label{eq: degrade}
   I_{d}\left( x,y \right)=PSF_{c}\left( x,y \right)\ast I_{s}\left( x,y \right)+\mathcal{N}.
\end{equation}

In our experiments, the pixel size is set to $1.2 um$ and the sensor resolution is 3000x4000, corresponding to a diagonal image height of 3 mm. For computational efficiency during validation, the sensor is divided into 15x20 blocks of 200x200 pixels each, with the assumption that the PSF within each block is spatially invariant.

\section{Joint Optimization Pipline}\label{sec: pip}
In traditional optical design software, the evaluation of image quality performance typically involves partial information about the PSF, such as spot diagram root mean square (RMS), wavefront RMS, Strehl ratio, and modulation transfer function (MTF) at specific frequencies. With the development of image post-processing techniques, the balance of PSF across different fields of view has become a more central factor in reconstructed image quality evaluation, while traditional optical optimization cannot evaluate and optimize the specific energy distribution of the PSF. Therefore, we construct an end-to-end joint optimization pipeline based on accurately computed aberration-degraded PSF and regard network as the whole PSF merit function for successive joint optimization.

\subsection{Image Reconstruction Network}
MIMO-UNet\cite{mimo} is a fast and accurate deblurring network designed with multi-scale features and residual blocks. Balancing computational efficiency with performance, MIMO-UNet is well-suited to provide an excellent overall evaluation of the PSF in our successive joint design.

As is shown in Fig. \ref{fig:joint}, we utilize the previous differentiable optical simulation model to calculate the accurate PSF for arbitrary 2D field on the image plane and construct degraded images using the method described in Sec. \ref{sec: degrade}. Unlike typical image quality issues like motion blur, degradation caused by optical aberrations is highly correlated with the FoV. To address the spatially varying degradation across different fields, we incorporate pixel-by-pixel relative positioning as prior information into the network, along with the three-channel image. Specifically, we normalize the Cartesian coordinates of all field pixels based on the sensor's diagonal resolution, concatenate them with the input image in the channel dimension, and feed them into the reconstruction network. The input channels of MIMO-UNet are modified to 5 to match the characteristics of the input data, while the rest of the network architecture remains unchanged.

\subsection{Loss Functions} \label{sec: loss}
In addition to evaluating image quality, optical design often requires constraints on design specifications and manufacturability. Therefore, beyond using neural network loss functions to enhance the quality of reconstructed images, we also introduce several common optical system constraints to ensure that the lens system parameters meet the necessary design requirements.

\textbf{Imaging Reconstruction Loss.} We reconstruct aberration-degraded images $I_{d}$ through an image reconstruction network to produce reconstructed images $I_{r}$. The quality of the restored images is assessed using the same multi-scale loss function as in MIMO-UNet
\begin{equation}\label{eq: net loss}
   L_{net}=\frac{1}{t_{k}} \sum_{k=1}^{K} \Vert I_{r,k} - I_{d,k} \Vert_1 + \lambda_{f}\Vert \mathcal{F}\left( I_{r,k} \right) - \mathcal{F}\left( I_{d,k} \right) \Vert_1, 
\end{equation}
where $K$, $t_{k}$, and $\mathcal{F}$ represent the number of levels, normalization factor and FFT, respectively. We set $K=3$, $\lambda_{f}=0.1$. $L_{net}$ represents the network's ability to decode scene information encoded by the PSF, serving as an overall evaluation of the PSF. This approach differs significantly from the typical merit functions used in traditional optical design software.

\textbf{Optical Losses.} 
The total optical length is typically a critical design parameter for lens modules. During the joint optimization process, we constrain the total length to not exceed a specified threshold $TTL_{max}$:
\begin{equation}\label{eq: ttl loss}
   L_{ttl}=\max \left( \sum_{j}^{S} d_{j}, TTL_{max}\right),
\end{equation}
where $S$ represents  the total surface number and $d_{j}$ denotes thickness of $j$-th surface.

For an optical system with a sensor image height $IH$ and  a maximum FoV, the effective focal length (EFFL) of the lens module, calculated  through paraxial optics, should be controlled around a specific target. The corresponding loss function is formulated as
\begin{equation}\label{eq: effl loss}
   L_{effl}= \Vert EFFL - \frac{IH}{2\tan\left( FoV/2 \right)} \Vert_1.
\end{equation}

In optical optimization, it is essential to control the spacing between adjacent surfaces to prevent surface self-intersection. We sample 256 points based on the surface aperture and measure the axial distance $\Delta z$ from these points to the adjacent surface. The surface gap loss function is defined as
\begin{equation}\label{eq: gap loss}
   L_{gap}= - \sum_{j}^{S} \sum_{i}^{256} \min \left( z_{ji}, \epsilon_{gap} \right),
\end{equation}
where $\epsilon_{gap}$ represents the minimum spacing threshold.

The RMS of the spot diagram is one of the most fundamental and widely used metrics for evaluating image quality. The spot diagram loss function can be expressed as
\begin{equation}\label{eq: spot loss}
   L_{spot}= \sqrt{\mathop{avg}\limits_{N}\left( \left( x_{wp} - x_{c} \right)^{2} + \left( y_{wp} - y_{c} \right)^{2} \right)},
\end{equation}
where $x$, $y$ are coordinates of rays on the image plane, $w$ and $p$ represent a specified ray sampled over various wavelengths and entrance pupil coordinates, respectively, $N$ is total number of the rays and $c$ denotes the chief ray of reference wavelength.

Distortion refers to the deviation of the actual image point from the ideal image point. Since the exact image plane is typically not located at the paraxial image plane, we trace a small-angle ray from the entrance pupil to the actual image plane and calculate the system's Distortion Focal Length (DFFL).
\begin{equation}\label{eq: dist loss}
   L_{dist}= \max \left( \lvert \frac{r_{c} - DFFL\tan(v)}{DFFL\tan(v)} \rvert, \epsilon_{dist} \right), \forall v \neq 0,
\end{equation}
where v is the field of view and $r_{c}$ denotes the image point of chief ray. Only absolute distortions exceeding the distortion tolerance threshold $\epsilon_{dist}$ are penalized.

All the aforementioned optical loss terms are combined to define an optical loss $L_{optic}$, which operates exclusively on the lens design parameters $\phi_{lens}$
\begin{equation}\label{eq: optical loss}
   L_{optic}= L_{spot} + \lambda_{t}L_{ttl} + \lambda_{f}L_{effl} + \lambda_{g}L_{gap} + \lambda_{d}L_{dist},
\end{equation}
where we set weighting factors $\lambda_{t}=0.5$, $\lambda_{f}=10$, $\lambda_{g}=3$,and $\lambda_{d}=5$ to balance the contributions of each optical constraint. $L_{optic}$ plays a pivotal role in the successive optimization process, constraining the optical design space within a physically feasible domain that meets product requirements. In implementation, $L_{optic}$ can independently serve as an evaluation function for optimizing the optical system, and it can also combined with the network reconstruction losses $L_{net}$ to define the joint loss
\begin{equation}\label{eq: total loss}
   L_{joint}= L_{net} + \lambda_{lens}L_{optic},
\end{equation}
where $\lambda_{lens}$ is set individually for each lens.

\begin{figure*}[ht]
    \centering
    \includegraphics[width=0.9\textwidth]{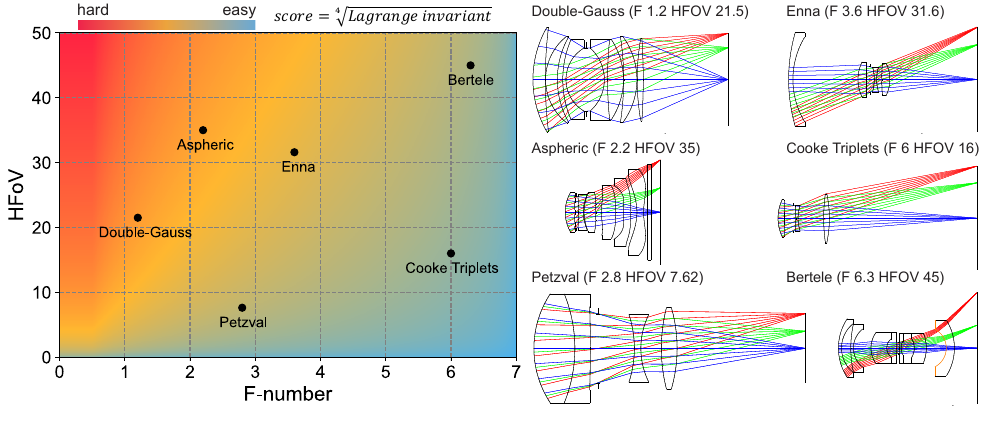}
    \caption{Overview of the lenses specifications and layouts in the experiments. The relative difficulty of each lens design is estimated based on the optical Lagrange invariant, $nyu$. All of these lenses are utilized for PSF validation in Sec. \ref{sec:psf valid}, with the top-left three lenses also being used for joint optimization in Sec. \ref{sec:joint}.}
    \label{fig:layout}
\end{figure*}

\section{Experiments and discussion}\label{experiments}
In this section, we validate the accuracy and robustness of the proposed method across various lenses, compare it with existing ray-based differentiable PSF calculation methods, and demonstrate the effectiveness of our joint optimization pipeline in several design examples. In Sec. \ref{sec:psf valid}, we first evaluate the accuracy of the initial value strategy for intersection points. Following this, we introduce the test lenses, then compare different PSF calculation methods, and analyze the performance of our differential operator. Finally, the details and results of our end-to-end design approach are presented in Sec. \ref{sec:joint}, which successively improves the quality of optical imaging systems.

\subsection{PSF Calculation Experiments}\label{sec:psf valid}

\textbf{Validation of the initial value strategy.} We uniformly sample 1,000 rays on the aperture plane and calculate their intersection points with the surface, as illustrated in Fig. \ref{fig:schematic}a. This calculation is performed using a simple estimation strategy (baseline) or the proposed initial value scheme, and the true value of intersection is obtained from non-gradient progressive approximation. As shown in Table \ref{table: intersection}, the results of both methods are similar for small field angles. However, as the FoV increases, our well-estimated initial values ensure accurate inner intersection points and stable Newton's method iterations, effectively avoiding the risk of converging to incorrect intersection points outside the aperture.

\begin{table}[!ht]
\caption{Performance of different initial value strategies for Newton's methods.}\label{table: intersection}
\renewcommand{\arraystretch}{1.3}
\centering
\resizebox{1.0\linewidth}{!}{
\begin{tabular}{c|cccccc}
\hline
FoV               & $20^{\circ}$  & $24^{\circ}$  & $28^{\circ}$  & $23^{\circ}$  & $36^{\circ}$  & $40^{\circ}$     \\ \hline
Baseline iters    & 5             & 722           & 691           & 1342          & 905           & 1030                       \\ \hline
Proposed iters    & 5             & 5             & 5             & 5             & 5             & 6                             \\ \hline
Baseline accuracy & $100 \%$      & $99.5 \%$     & $98.2 \%$     & $96.4 \%$     & $93.9 \%$     & $92.1 \%$              \\ \hline
Proposed accuracy & $100 \%$      & $100 \%$      & $100 \%$      & $100 \%$      & $100 \%$      & $100 \%$                \\ \hline
\end{tabular}
}
\end{table}

\textbf{Details of the Test Lenses.} We select five classical lenses from the ZEBASE\cite{zebase} along with a advanced aspheric lens, as shown in in Fig\ref{fig:layout}, to assess the accuracy and robustness of PSF calculation results across diverse lens scenarios. These lenses encompass a wide range of structures, f-numbers, FoV, and element counts. To facilitate comparisons, the image height of all lenses is scaled to 3mm.

For each lens, PSFs is evaluated at three design wavelengths and across 10 field points, ranging from on-axis to the maximum field of view. The entrance pupil is sampled at 129x129, and the image plane PSF is computed with a pixel size of $0.6 um$ and a grid size of 63x63 pixels. The Huygens PSF (ground truth) data is obtained through Zemax, while the other methods (coherent PSF, geometric PSF\cite{Yang_2024}, and Gaussian PSF\cite{cote2023differentiable}) are calculated based on the same accurate ray tracing data.

\textbf{Verification of PSF Calculation Accuracy and Stability.}
\begin{figure}[ht]
    \includegraphics[width=0.48\textwidth]{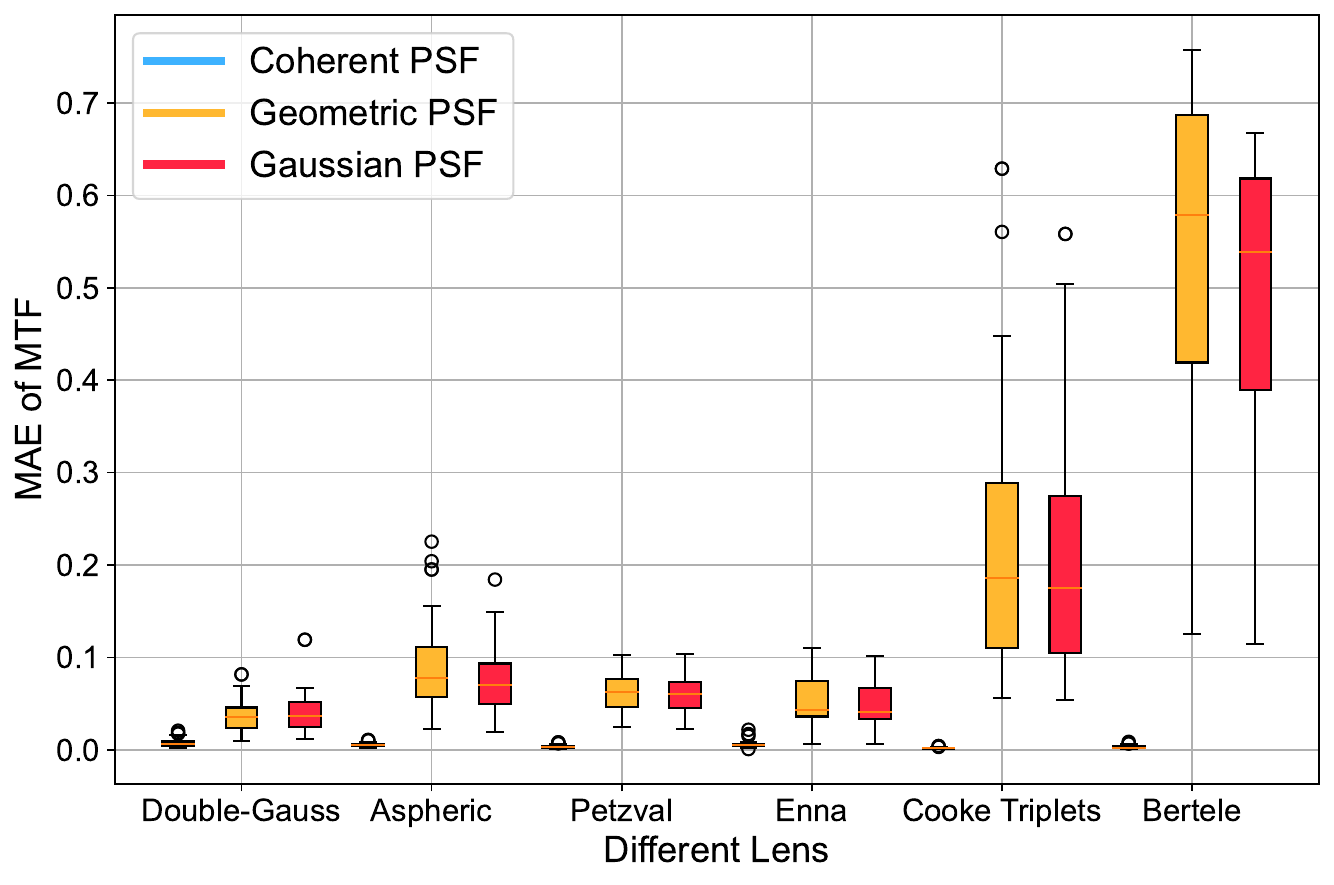}
    \caption{MAE of MTF for different PSF calculation methods across various lens systems. The central line within each box represents the median, while the length of the box indicates the interquartile range. The black circle denotes an outlier. Each set of three boxes corresponds to one of the PSF calculation methods.}
    \label{fig:MTF}
\end{figure}
We calculate the MTF of each single-wavelength PSF using Fourier transforms and compare their accuracy to the ground truth (MTF of Zemax Huygens PSF) in the sagittal and tangential directions. As shown in Fig \ref{fig:MTF}, the results obtained through the coherent PSF method are generally superior, with the mean absolute error (MAE) of the MTF being nearly zero. In contrast, the other two incoherent methods exhibit some computational errors.

\begin{figure}[ht]
    \centering
    \includegraphics[width=0.5\textwidth]{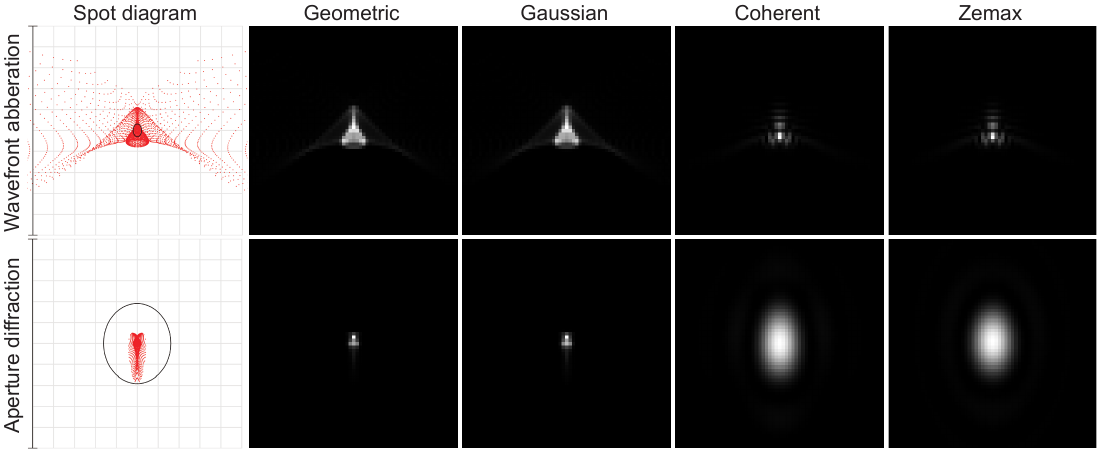}
    \caption{Visualization of different PSF calculation results. The Airy disks are outlined by black elliptical lines in the spot diagrams. The results in the first row, which is on Double-Gauss lens, illustrates that PSFs under wavefront aberrations can cause destructive interference within uniformly distributed spots. The second row, featuring Bertele lens (F/6.3, corresponding to an Airy disk size of approximately 7 pixels), shows that the aperture diffraction effect result in PSFs with greater spread compared to the converged spots distribution.}
    \label{fig:psf}
\end{figure}

To illustrate the differences between the methods more intuitively, we present single-wavelength PSFs calculated by various methods in Fig \ref{fig:psf}. For the Double-Gauss lens, the coherent PSF accurately captures the energy ripples caused by wave aberrations, while the incoherent methods only describe the overall energy distribution. For the large f-number Bertele lens, although the ray spot diagram converges well, exit pupil diffraction effect causes significant energy dispersion, leading to noticeable deviations between the incoherent methods and the ground truth.

These experimental comparisons highlight that incoherent PSF calculation methods may exhibit substantial deviations in the presence of large wave aberrations and pronounced aperture diffraction effects. In contrast, our PSF calculation method is more general, it could maintain accuracy and stability comparable to Zemax, ensuring that the computational precision required for complex lens joint designs meets commercial application standards.

\begin{figure}[ht]
\subfloat[Memory comparison]{\includegraphics[width=1.7 in]{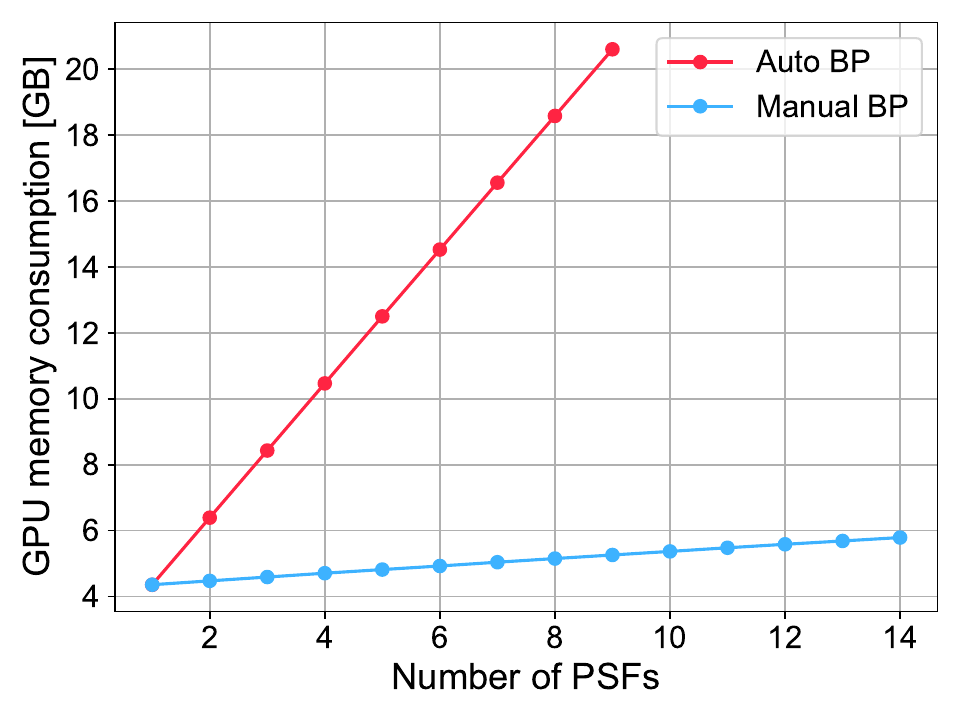}%
\label{Memory comparison}}
\subfloat[Timing comparison]{\includegraphics[width=1.7 in]{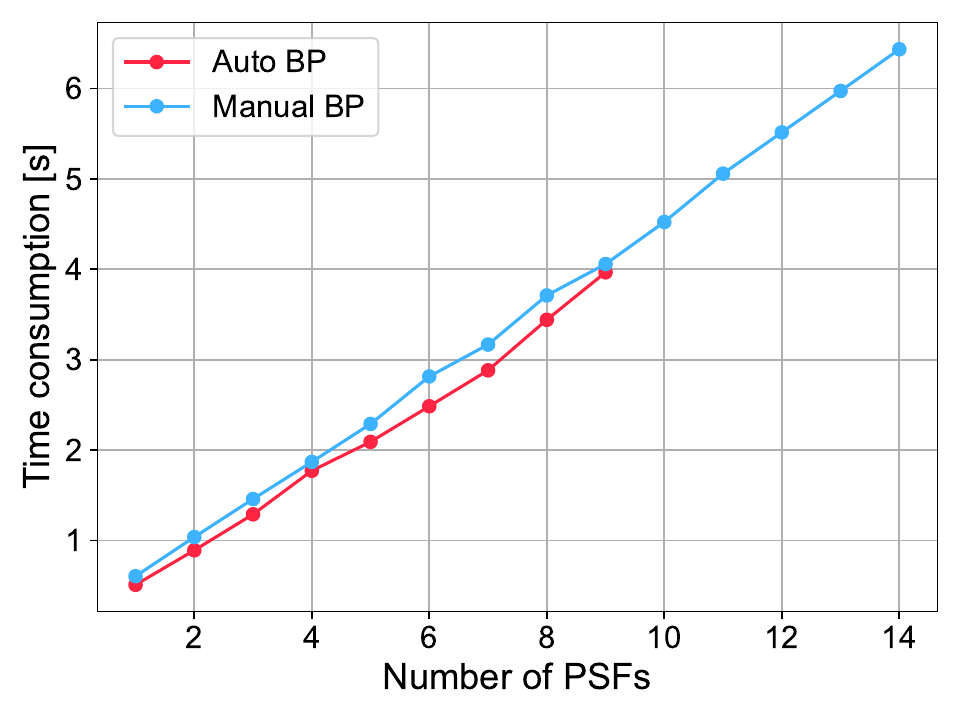}%
\label{Timing comparison}}
\caption{The comparison of memory and time costs between traditional automatic BP and our operator-based manual BP is shown with respect to the number of PSFs. Due to excessive memory requirements, the conventional method cannot handle more than 9 PSFs. In contrast, the differential operator approach effectively manages larger numbers of PSFs while maintaining low memory consumption.}
\label{fig:operator}
\end{figure}

\textbf{The performance of differential operator.}
To validate the performance of the proposed differential operator against automatic differentiation functions, we used a dummy PSF merit function in Fig. \ref{fig:operator} to compare the memory and time costs of both methods, focusing on the number of sampling PSFs under the sampling density defined in this section.

Compared to basic automatic BP, the operator-based BP takes only slightly more time but reduces memory usage by approximately 18.4 times per PSF. This significant reduction in memory requirements supports scaling up the number of sampled PSFs and enables practical application in end-to-end computational imaging design tasks by efficiently decoupling ray and grid dimensions in coherent PSF calculations.

\subsection{Joint Optimization Experiments}\label{sec:joint}
Lens design generally becomes more challenging with a larger HFoV or a smaller F-number. We selected three relatively complex lenses (Double-Gauss, Enna, and Aspheric) from our test lenses for joint optimization evaluation, as detailed in Sec.\ref{sec: pip}. The goal is to successively achieve the highest possible image quality while ensuring that the optical system meets design requirements and manufacturability constraints.

\textbf{Lens specifications.} In addition to the curvatures $c$ and thickness $d$ of spherical surfaces, the optical optimization variables for aspherical surfaces also include conic constant $k$, and aspheric coefficients $a_{i}$. To mitigate the nonlinearity in optical optimization, we balanced the learning rates of these variables relative to their influence on the effective focal length (EFFL). Specifically, for a given learning rate $\eta$ for the lens, the learning rate for $c$,$d$,$k$ and $a_{i}$ are set to $\eta/f$, $\eta f$, $\eta$, and $\eta/f^{i-1}$, respectively. This approach is analogous to scaling the lens to EFFL of 1 for optimization, similar to the method described in \cite{Cote:21}. 

\begin{table}[ht]
\renewcommand{\arraystretch}{1.3}
\caption{Design Specifications for three lenses.}\label{table: specifications}
\centering
\resizebox{1.0\linewidth}{!}{
\begin{tabular}{c|cccc|cc}
\hline
Lens         & FoV & F-number & TTL & EFFL & $\epsilon_{gap}$                     & $\epsilon_{dist}$             \\ \hline
Double-Gauss & $43^{\circ}$    & 1.2      & {12.39 mm}    & {7.616 mm}     & \multirow{3}{*}{0.02 mm}  & \multirow{3}{*}{0.5\%} \\
Enna         & $63.2^{\circ}$  & 3.5      & {10.74 mm}    & {4.876 mm}     &                           &                        \\
Aspheric     & $70^{\circ}$    & 2.2      & {5.28 mm}     & {4.285 mm}     &                           &                        \\ \hline
\end{tabular}
}
\end{table}

In Table \ref{table: specifications}, we outline the optical constraints for lens design tasks: the total length should not exceed that of the original design, the effective focal length should remain approximately constant, lens distortion and point spread dispersion should be minimal, and the minimum thickness interval should be constrained to ensure manufacturability. The optical weights $\lambda_{optic}$ are set to 1 for all these design tasks.

\textbf{Network training.} We use the DIV2K\cite{div2k} dataset, which contains 800 images at 2K resolution, for evaluation. During network training, the data is randomly cropped to a patch size of 256x256, with a batch size of 16 for each iteration. We employed the Adam optimizer with an initial learning rate of $10^{-4}$ and a multi-step scheduler that halves the learning rate every 200 epochs, training for a total of 1000 epochs. The entire model requires approximately 30 hours to train on one NVIDIA RTX 3090 GPU. After training, we use 63 images from the DIV8K\cite{div8k} dataset, center-cropped to a resolution of 3000x4000, to construct the input data, as described in Sec.\ref{sec: degrade}, and then evaluate image quality of output results.

\textbf{Qualitative evaluation.}
We report the quality of our recovered images, both with joint design optimization and with a fixed lens, in Table \ref{table: quality}, using the averaged PSNR, SSIM, and LPIPS metrics. Additionally, we provide the optical degraded image quality metrics to further demonstrate the improvements achieved through joint optimization of these lenses. All successively joint-optimized lenses outperforms the original designs in terms of both degraded and reconstructed image quality metrics.
\begin{table}[ht]
\renewcommand{\arraystretch}{1.3}
\caption{Performance comparison of the original design and joint design.}\label{table: quality}
\centering
\resizebox{1.0\linewidth}{!}{
\begin{tabular}{c|ccc|ccc}
\hline
\multirow{2}{*}{Experiment} & \multicolumn{3}{c|}{Imaging} & \multicolumn{3}{c}{Recovery}            \\ \cline{2-7} 
                            & ${PSNR\uparrow}$  & ${SSIM\uparrow}$   & ${LPIPS\downarrow}$  & ${PSNR\uparrow}$  & ${SSIM\uparrow}$   & ${LPIPS\downarrow}$                       \\ \hline
Double-Gauss                & 26.43 & 0.5918 & 0.3030 & 35.24 & 0.9357 & 0.0691                      \\
Double-Gauss Joint          & 27.48 & 0.6287 & 0.2449 & 36.86 & 0.9478 & 0.0484                      \\ \hline
Enna                        & 25.56 & 0.5332 & 0.4091 & 33.02 & 0.9049 & 0.1139                      \\
Enna Joint                  & 27.88 & 0.6334 & 0.2424 & 36.43 & 0.9468 & 0.0537                      \\ \hline
Aspheric                    & 29.25 & 0.6735 & 0.1782 & 38.07 & 0.9608 & 0.0385                      \\
Aspheric Joint              & 29.68 & 0.6818 & 0.1691 & 38.64 & 0.9636 & 0.0324                      \\ \hline
\end{tabular}
}
\end{table}

\begin{figure*}[ht]
    \centering
    \includegraphics[width=0.8\textwidth]{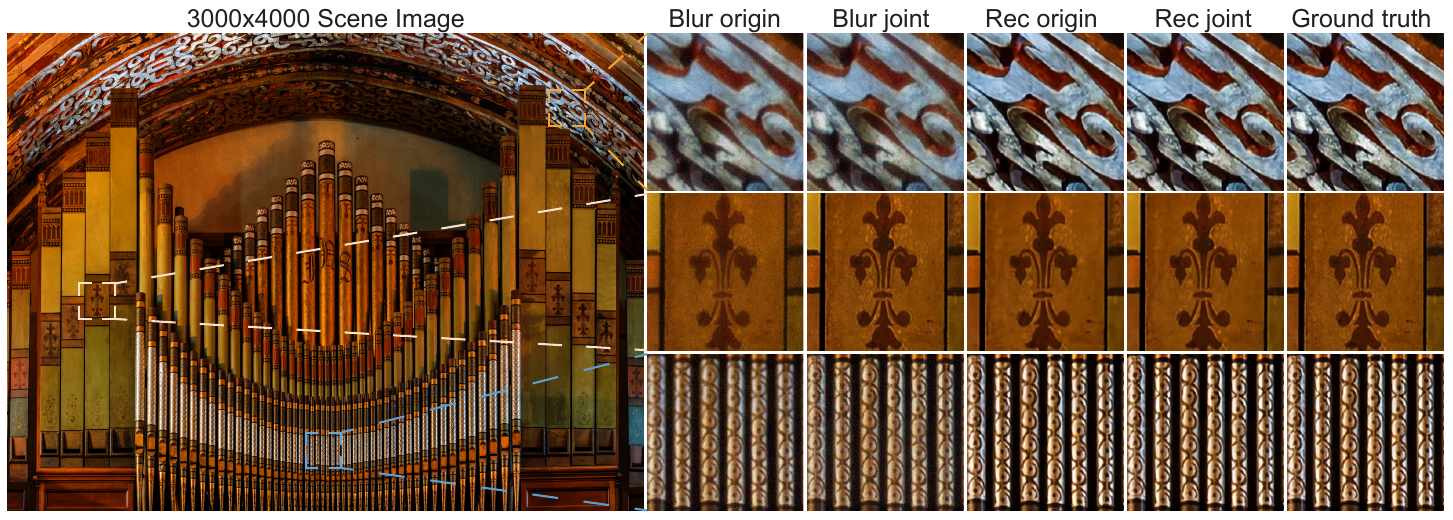}
    \caption{Visual comparison of the blurred and reconstructed images for Double-Gauss lens, with and without joint design. The magnified patches on the right correspond to the highlighted areas in the dashed boxes on the left..}
    \label{fig:visual}
\end{figure*}

The restoration network, with field information encoding, can adaptively perceive the varying difficulty of PSF restoration across different fields, eliminating the need for a consistent PSF across the entire image. As a result, our joint optimization pipeline allows the optical design to successively explore PSFs that yield better restored image quality metrics over a broader solution space. Fig. \ref{fig:visual} visualizes the blurred and reconstructed image patches for the Double-Gauss lens, with and without joint design, to subjectively corroborate the superior image quality metrics presented in Table \ref{table: quality}. 

\begin{figure*}[ht]
    \centering
    \includegraphics[width=0.8\textwidth]{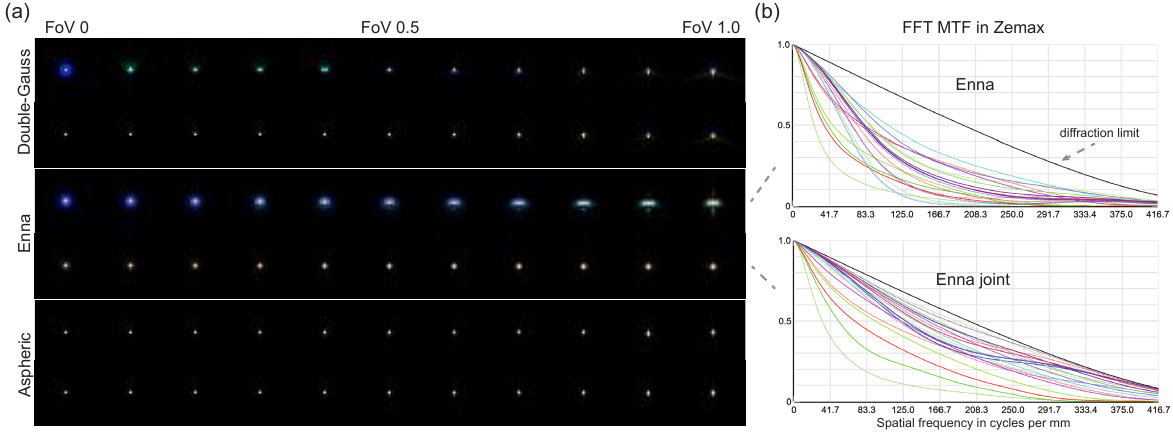}
    \caption{(a) compares the PSF map of different lenses before (upper) and after (lower) joint design. Note that for visualization purposes, each PSF is normalized according to the maximum value in three channels. (b) presents the FFT MTF of Enna's PSFs, with the topmost black line representing the diffraction limit.}
    \label{fig:line psf}
\end{figure*}

Furthermore, Fig. \ref{fig:line psf} offers an optical perspective on the changes in PSF and MTF for the lenses before and after joint optimization. The results show that, after successive joint optimization, the PSFs of all three lenses become more focused, indicating that their MTFs are closer to the diffraction limit.

\begin{figure*}[!ht]
    \centering
    \includegraphics[width=0.8\textwidth]{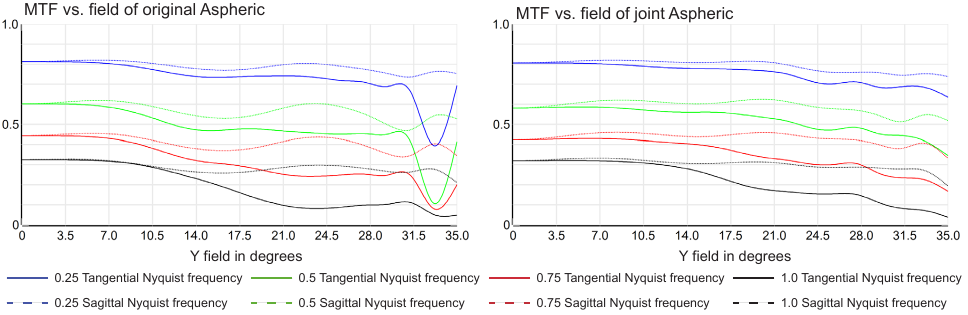}
    \caption{Zemax MTF vs. field analysis comparison between original (left) and joint designed (right) Aspheric. Blue, green, red, and black lines represent 0.25, 0.5, 0.75, and 1.0 Nyquist frequencies, respectively. Solid lines indicate the tangential direction, while dashed lines indicate the sagittal direction.}
    \label{fig:mtf_field}
\end{figure*}

Unlike traditional optical tasks that evaluate a few discrete fields, our random field training approach more effectively detects and balances wavefront aberration changes across different fields. As illustrated in Fig. \ref{fig:mtf_field}, our Aspheric lens, originally well optimized using Zemax for 10 discrete fields, exhibits significant degradation around the $33^{\circ}$ field while maintaining good mtf in adjacent design fields ($31.1^{\circ}$ and $35^{\circ}$). The joint optimization pipeline successively corrects this deficiency, leading to improved optical image quality and realizing field-level PSF control.

Additionally, as shown in  Fig.\ref{fig:line psf} and Fig.\ref{fig:mtf_field}, there is a noticeable trend in the jointly optimized network to mitigate optical degradation by boosting the lower MTF values across the Nyquist frequency. This improvement implies that achieving a higher integrated EMTF depends on the signal-to-noise ratio (SNR): $\frac{MTF^{2}}{MTF^{2}+1/SNR}$.

\subsection{Ablation study}
A comprehensive ablation study is performed to verify that every step in our method is necessary, as shown in Table \ref{table: specifications after design} and Table \ref{table: ablations}. Here we show the ablation study on the proposed joint optimization pipeline, where we employ a basic MIMO-UNet model and delete some key components for evaluations. The recovery column shows the Lens+Net results and the imaging column shows only the Lens imaging simulation results. For joint optimization training, the results show the necessity of each module. 

\begin{table*}[ht]
\renewcommand{\arraystretch}{1.3}
\caption{Lens extrinsic parameters and situations after optimization.}\label{table: specifications after design}
\centering
\resizebox{0.9\linewidth}{!}{
\begin{tabular}{c|ccccc}
\hline
Lens                            & TTL constrain & $\Delta$EFFL        & no self-intersection  & Mean spots RMS     & Max distortion      \\ \hline
Double-Gauss joint        & $\checkmark$  & {-0.31 $\mu$m}          & $\checkmark$          & {14.2836 $\mu$m}       & $2.5093\%$            \\
Double-Gauss w/o $L_{optic}$  & $\checkmark$  & \textbf{-45.06 $\mu$m}  & \XSolidBrush          &  -                 &  -                       \\ \hline
Enna joint                & $\checkmark$  & {-11.72 $\mu$m}         & $\checkmark$          & {3.4702 $\mu$m}        & $2.1787\%$               \\
Enna w/o $L_{optic}$          & $\checkmark$  & \textbf{-179.01 $\mu$m} & $\checkmark$          & {3.0425 $\mu$m}        & \textbf{5.0680\%}           \\ \hline
Aspheric joint            & $\checkmark$  & {-2.08 $\mu$m}          & $\checkmark$          & {1.6792 $\mu$m}        & $3.4889\%$             \\ 
Aspheric w/o $L_{optic}$      & $\checkmark$  & \textbf{-74.90 $\mu$m}  & $\checkmark$          & \textbf{1.7096 $\mu$m} & \textbf{4.1113\%}       \\ \hline
\end{tabular}
}
\end{table*}

\textbf{Ablation on optical constrains.} Without optical constraints $L_{optic}$, joint optimization may cause the lens design to deviate from the specified requirements, leading to significant deviations in EFFL, surface self-intersections, and excessive distortions, ultimately resulting in impractical designs.

\begin{table*}[ht]
\renewcommand{\arraystretch}{1.3}
\caption{Ablation study on the joint optimization of the Aspheric lens.}\label{table: ablations}
\centering
\resizebox{0.75\linewidth}{!}{
\begin{tabular}{c|ccc|ccc}
\hline
\multirow{2}{*}{Experiment}           & \multicolumn{3}{c|}{Imaging} & \multicolumn{3}{c}{Recovery}            \\ \cline{2-7} 
                                      & ${PSNR\uparrow}$  & ${SSIM\uparrow}$   &${ LPIPS\downarrow}$  & ${PSNR\uparrow}$  & ${SSIM\uparrow}$   & ${LPIPS\downarrow}$    \\ \hline
Complete methodology                  & 29.68         & 0.6818         & 0.1691           & 38.64         & 0.9636         & 0.0324     \\
Fixed $\phi_{lens}$                   & 29.25         & 0.6735         & 0.1782           & 38.07         & 0.9608         & 0.0385              \\ 
No field information                  & 29.74         & 0.6827         & 0.1688           & 38.11         & 0.9610         & 0.0356              \\ 
$L_{net}$ detached from $\phi_{lens}$ & 28.41         & 0.6542         & 0.1948           & 37.36         & 0.9534         & 0.0435              \\ \hline
\end{tabular}
}
\end{table*}

\textbf{Ablation on field information.} The interesting observation in this situation is that when field information is removed, the unsatisfied evaluations in the deep recovery model lead to a lager $L_{net}$, forcing the lens to sacrifice some optical constraints $L_{optic}$ (e.g., with distortion worsening from 3.4889\% to 3.6744\%) in pursuit of slightly higher image quality. However, the complete methodology allows for better control of optical constraints while simultaneously achieving superior combined recovery image clarity. This result emphasis the necessity of incorporating field information for the restoration network to work as a powerful PSF merit function across FoV.

\textbf{Ablation of network merit function.} When $L_{net}$ is not used as an overall PSF merit function and only $L_{optic}$ is applied for joint optimization, the lens’s optical performance cannot be further enhanced, thereby limiting the final imaging quality.

\section{Conclusion}\label{conclution}
Previous works in ray-based differentiable PSF calculation methods ignore the wave nature of light, so they are not suitable for arbitrary cases, here we establish a novel differentiable optical simulation to analyze imaging systems constrained by wavefront aberrations and diffraction effects. The initial guess strategy for Newton's method ensures the confidence of intersection points for high aspherics. While maintaining the accuracy of results, our differentiable coherent PSF operator efficiently saves memory occupancy during calculations by manually back-propagating gradients. This allows for accurate PSF computations to be lightweight and widely applicable in any joint design tasks. Furthermore, we propose a joint optimization pipeline with field information to solve spatially variant optical degradation. Experiments verified that our method successively achieve superior optical performance and reconstructed image quality on well designed systems. Along with the release of code, we hope this work will enhance the precision of optical simulations and further unlock the potential of advanced compact lens in joint design tasks.

\section*{Acknowledgments}
We thank Meijuan Bian from the facility platform of optical engineering of Zhejiang University for instrument support.

\bibliographystyle{IEEEtranS}
\bibliography{ref}

\begin{thebibliography}{10}
\providecommand{\url}[1]{#1}
\csname url@samestyle\endcsname
\providecommand{\newblock}{\relax}
\providecommand{\bibinfo}[2]{#2}
\providecommand{\BIBentrySTDinterwordspacing}{\spaceskip=0pt\relax}
\providecommand{\BIBentryALTinterwordstretchfactor}{4}
\providecommand{\BIBentryALTinterwordspacing}{\spaceskip=\fontdimen2\font plus
\BIBentryALTinterwordstretchfactor\fontdimen3\font minus \fontdimen4\font\relax}
\providecommand{\BIBforeignlanguage}[2]{{%
\expandafter\ifx\csname l@#1\endcsname\relax
\typeout{** WARNING: IEEEtranS.bst: No hyphenation pattern has been}%
\typeout{** loaded for the language `#1'. Using the pattern for}%
\typeout{** the default language instead.}%
\else
\language=\csname l@#1\endcsname
\fi
#2}}
\providecommand{\BIBdecl}{\relax}
\BIBdecl

\bibitem{tensorflow}
\BIBentryALTinterwordspacing
M.~Abadi, A.~Agarwal, P.~Barham, E.~Brevdo, Z.~Chen, C.~Citro, G.~S. Corrado, A.~Davis, J.~Dean, M.~Devin, S.~Ghemawat, I.~Goodfellow, A.~Harp, G.~Irving, M.~Isard, Y.~Jia, R.~Jozefowicz, L.~Kaiser, M.~Kudlur, J.~Levenberg, D.~Mane, R.~Monga, S.~Moore, D.~Murray, C.~Olah, M.~Schuster, J.~Shlens, B.~Steiner, I.~Sutskever, K.~Talwar, P.~Tucker, V.~Vanhoucke, V.~Vasudevan, F.~Viegas, O.~Vinyals, P.~Warden, M.~Wattenberg, M.~Wicke, Y.~Yu, and X.~Zheng, ``Tensorflow: Large-scale machine learning on heterogeneous distributed systems,'' 2016. [Online]. Available: \url{https://arxiv.org/abs/1603.04467}
\BIBentrySTDinterwordspacing

\bibitem{div2k}
E.~Agustsson and R.~Timofte, ``Ntire 2017 challenge on single image super-resolution: Dataset and study,'' in \emph{The IEEE Conference on Computer Vision and Pattern Recognition (CVPR) Workshops}, July 2017.

\bibitem{DeepOpticsDepth}
J.~Chang and G.~Wetzstein, ``Deep optics for monocular depth estimation and 3d object detection,'' in \emph{2019 IEEE/CVF International Conference on Computer Vision (ICCV)}, 2019, pp. 10\,192--10\,201.

\bibitem{Chen_2021}
\BIBentryALTinterwordspacing
S.~Chen, H.~Feng, D.~Pan, Z.~Xu, Q.~Li, and Y.~Chen, ``Optical aberrations correction in postprocessing using imaging simulation,'' \emph{ACM Transactions on Graphics}, vol.~40, no.~5, p. 1–15, Sep. 2021. [Online]. Available: \url{http://dx.doi.org/10.1145/3474088}
\BIBentrySTDinterwordspacing

\bibitem{Chen_2023}
\BIBentryALTinterwordspacing
S.~Chen, T.~Lin, H.~Feng, Z.~Xu, Q.~Li, and Y.~Chen, ``Computational optics for mobile terminals in mass production,'' \emph{IEEE Transactions on Pattern Analysis and Machine Intelligence}, vol.~45, no.~4, p. 4245–4259, Apr. 2023. [Online]. Available: \url{http://dx.doi.org/10.1109/TPAMI.2022.3200725}
\BIBentrySTDinterwordspacing

\bibitem{mimo}
\BIBentryALTinterwordspacing
S.-J. Cho, S.-W. Ji, J.-P. Hong, S.-W. Jung, and S.-J. Ko, ``Rethinking coarse-to-fine approach in single image deblurring,'' 2021. [Online]. Available: \url{https://arxiv.org/abs/2108.05054}
\BIBentrySTDinterwordspacing

\bibitem{Chung:19}
\BIBentryALTinterwordspacing
J.~Chung, G.~W. Martinez, K.~C. Lencioni, S.~R. Sadda, and C.~Yang, ``Computational aberration compensation by coded-aperture-based correction of aberration obtained from optical fourier coding and blur estimation,'' \emph{Optica}, vol.~6, no.~5, pp. 647--661, May 2019. [Online]. Available: \url{https://opg.optica.org/optica/abstract.cfm?URI=optica-6-5-647}
\BIBentrySTDinterwordspacing

\bibitem{Cote:21}
\BIBentryALTinterwordspacing
G.~C\^{o}t\'{e}, J.-F. Lalonde, and S.~Thibault, ``Deep learning-enabled framework for automatic lens design starting point generation,'' \emph{Opt. Express}, vol.~29, no.~3, pp. 3841--3854, Feb 2021. [Online]. Available: \url{https://opg.optica.org/oe/abstract.cfm?URI=oe-29-3-3841}
\BIBentrySTDinterwordspacing

\bibitem{cote2023differentiable}
G.~C{\^o}t{\'e}, F.~Mannan, S.~Thibault, J.-F. Lalonde, and F.~Heide, ``The differentiable lens: Compound lens search over glass surfaces and materials for object detection,'' in \emph{Proceedings of the IEEE/CVF Conference on Computer Vision and Pattern Recognition (CVPR)}, June 2023.

\bibitem{Dun:20}
\BIBentryALTinterwordspacing
X.~Dun, H.~Ikoma, G.~Wetzstein, Z.~Wang, X.~Cheng, and Y.~Peng, ``Learned rotationally symmetric diffractive achromat for full-spectrum computational imaging,'' \emph{Optica}, vol.~7, no.~8, pp. 913--922, Aug 2020. [Online]. Available: \url{https://opg.optica.org/optica/abstract.cfm?URI=optica-7-8-913}
\BIBentrySTDinterwordspacing

\bibitem{eboli2022fasttwostepblindoptical}
\BIBentryALTinterwordspacing
T.~Eboli, J.-M. Morel, and G.~Facciolo, ``Fast two-step blind optical aberration correction,'' 2022. [Online]. Available: \url{https://arxiv.org/abs/2208.00950}
\BIBentrySTDinterwordspacing

\bibitem{gao2024globalsearchopticsautomatically}
\BIBentryALTinterwordspacing
Y.~Gao, Q.~Jiang, S.~Gao, L.~Sun, K.~Yang, and K.~Wang, ``Global search optics: Automatically exploring optimal solutions to compact computational imaging systems,'' 2024. [Online]. Available: \url{https://arxiv.org/abs/2404.19201}
\BIBentrySTDinterwordspacing

\bibitem{div8k}
S.~Gu, A.~Lugmayr, M.~Danelljan, M.~Fritsche, J.~Lamour, and R.~Timofte, ``Div8k: Diverse 8k resolution image dataset,'' in \emph{2019 IEEE/CVF International Conference on Computer Vision Workshop (ICCVW)}, 2019, pp. 3512--3516.

\bibitem{Hale:21}
\BIBentryALTinterwordspacing
A.~Hal\'{e}, P.~Trouv\'{e}-Peloux, and J.-B. Volatier, ``End-to-end sensor and neural network design using differential ray tracing,'' \emph{Opt. Express}, vol.~29, no.~21, pp. 34\,748--34\,761, Oct 2021. [Online]. Available: \url{https://opg.optica.org/oe/abstract.cfm?URI=oe-29-21-34748}
\BIBentrySTDinterwordspacing

\bibitem{zebase}
M.~R. Hensley, E.~Hassenplug, R.~McPhail, and Y.~Leung, ``Zebase: an open-source relational database for zebrafish laboratories,'' \emph{Zebrafish}, vol.~9, pp. 44--49, 2012.

\bibitem{li2021universalflexibleopticalaberration}
\BIBentryALTinterwordspacing
X.~Li, J.~Suo, W.~Zhang, X.~Yuan, and Q.~Dai, ``Universal and flexible optical aberration correction using deep-prior based deconvolution,'' 2021. [Online]. Available: \url{https://arxiv.org/abs/2104.03078}
\BIBentrySTDinterwordspacing

\bibitem{Li:21}
\BIBentryALTinterwordspacing
Z.~Li, Q.~Hou, Z.~Wang, F.~Tan, J.~Liu, and W.~Zhang, ``End-to-end learned single lens design using fast differentiable ray tracing,'' \emph{Opt. Lett.}, vol.~46, no.~21, pp. 5453--5456, Nov 2021. [Online]. Available: \url{https://opg.optica.org/ol/abstract.cfm?URI=ol-46-21-5453}
\BIBentrySTDinterwordspacing

\bibitem{lin2022non}
T.~Lin, S.~Chen, H.~Feng, Z.~Xu, Q.~Li, and Y.~Chen, ``Non-blind optical degradation correction via frequency self-adaptive and finetune tactics,'' \emph{Optics Express}, vol.~30, no.~13, pp. 23\,485--23\,498, 2022.

\bibitem{DeepOpticsHDR}
C.~Metzler, H.~Ikoma, Y.~Peng, and G.~Wetzstein, ``Deep optics for single-shot high-dynamic-range imaging,'' in \emph{Proc. CVPR}, 2020.

\bibitem{nie2023freeform}
Y.~Nie, J.~Zhang, R.~Su, and H.~Ottevaere, ``Freeform optical system design with differentiable three-dimensional ray tracing and unsupervised learning,'' \emph{Optics Express}, vol.~31, no.~5, pp. 7450--7465, 2023.

\bibitem{pytorch}
\BIBentryALTinterwordspacing
A.~Paszke, S.~Gross, F.~Massa, A.~Lerer, J.~Bradbury, G.~Chanan, T.~Killeen, Z.~Lin, N.~Gimelshein, L.~Antiga, A.~Desmaison, A.~Köpf, E.~Yang, Z.~DeVito, M.~Raison, A.~Tejani, S.~Chilamkurthy, B.~Steiner, L.~Fang, J.~Bai, and S.~Chintala, ``Pytorch: An imperative style, high-performance deep learning library,'' 2019. [Online]. Available: \url{https://arxiv.org/abs/1912.01703}
\BIBentrySTDinterwordspacing

\bibitem{EndToEndCam}
\BIBentryALTinterwordspacing
V.~Sitzmann, S.~Diamond, Y.~Peng, X.~Dun, S.~Boyd, W.~Heidrich, F.~Heide, and G.~Wetzstein, ``End-to-end optimization of optics and image processing for achromatic extended depth of field and super-resolution imaging,'' \emph{ACM Trans. Graph.}, vol.~37, no.~4, Jul. 2018. [Online]. Available: \url{https://doi.org/10.1145/3197517.3201333}
\BIBentrySTDinterwordspacing

\bibitem{LearnedOpticHDR}
Q.~Sun, E.~Tseng, Q.~Fu, W.~Heidrich, and F.~Heide, ``Learning rank-1 diffractive optics for single-shot high dynamic range imaging,'' in \emph{The IEEE Conference on Computer Vision and Pattern Recognition (CVPR)}, June 2020.

\bibitem{Sun2021DiffLens}
Q.~Sun, C.~Wang, F.~Qiang, D.~Xiong, and H.~Wolfgang, ``End-to-end complex lens design with differentiable ray tracing,'' \emph{ACM Transactions on Graphics (TOG)}, vol.~40, no.~4, 2021.

\bibitem{Tseng2021DeepCompoundOptics}
E.~Tseng, A.~Mosleh, F.~Mannan, K.~St-Arnaud, A.~Sharma, Y.~Peng, A.~Braun, D.~Nowrouzezahrai, J.-F. Lalonde, and F.~Heide, ``Differentiable compound optics and processing pipeline optimization for end-to-end camera design,'' \emph{ACM Transactions on Graphics (TOG)}, vol.~40, no.~2, 2021.

\bibitem{dO}
C.~Wang, N.~Chen, and W.~Heidrich, ``do: A differentiable engine for deep lens design of computational imaging systems,'' \emph{IEEE Transactions on Computational Imaging}, vol.~8, pp. 905--916, 2022.

\bibitem{Yang_2024}
\BIBentryALTinterwordspacing
X.~Yang, Q.~Fu, and W.~Heidrich, ``Curriculum learning for ab initio deep learned refractive optics,'' \emph{Nature Communications}, vol.~15, no.~1, Aug. 2024. [Online]. Available: \url{http://dx.doi.org/10.1038/s41467-024-50835-7}
\BIBentrySTDinterwordspacing

\bibitem{zhou2024revealing}
J.~Zhou, S.~Chen, Z.~Ren, W.~Zhang, J.~Yan, H.~Feng, Q.~Li, and Y.~Chen, ``Revealing the preference for correcting separated aberrations in joint optic-image design,'' \emph{Optics and Lasers in Engineering}, vol. 178, p. 108220, 2024.

\end{thebibliography}

\vfill

\end{document}